\documentclass[letterpaper, 10 pt, conference]{ieeeconf}
\IEEEoverridecommandlockouts
\usepackage{cite}
\usepackage{amsmath,amssymb,amsfonts,amsthm}
\usepackage{graphicx}
\usepackage{textcomp}
\usepackage{xcolor}
\usepackage{url}
\usepackage{tablefootnote}

\usepackage{algorithm}
\usepackage{algpseudocode}
\usepackage{subcaption}
\usepackage{multirow}

\usepackage{booktabs}

\newcommand{\modelname}{\text{InterSim}}
\newcommand{\modelnamespace}{\text{InterSim} }

\title{\LARGE \bf
InterSim: Interactive Traffic Simulation via Explicit Relation Modeling
}

\author{Qiao Sun$^{1}$, Xin Huang$^{2}$, Brian C. Williams$^{2}$, and Hang Zhao$^{1*}$ 
 \thanks{$^{1}$IIIS, Tsinghua University
\texttt{alan.qiao.sun@gmail.com}
 }
\thanks{$^{2}$CSAIL, Massachusetts Institute of Technology}
\thanks{$^{*}$Corresponding at \texttt{hangzhao@mail.tsinghua.edu.cn}}
}

\begin{document}


\maketitle
 
\begin{abstract}
Interactive traffic simulation is crucial to autonomous driving systems by enabling testing for planners in a more scalable and safe way compared to real-world road testing. Existing approaches learn an agent model from large-scale driving data to simulate realistic traffic scenarios, yet it remains an open question to produce consistent and diverse multi-agent interactive behaviors in crowded scenes. In this work, we present InterSim, an interactive traffic simulator for testing autonomous driving planners. Given a test plan trajectory from the ego agent, InterSim reasons about the interaction relations between the agents in the scene and generates realistic trajectories for each environment agent that are consistent with the relations. We train and validate our model on a large-scale interactive driving dataset. Experiment results show that InterSim achieves better simulation realism and reactivity in two simulation tasks compared to a state-of-the-art learning-based traffic simulator. 

\end{abstract}

\section{Introduction}
With the recent development of autonomous driving technologies, traffic simulation has played an important role in enabling testing the planner system on a large scale. Compared to real-world road testing, simulation offers a more time and resource efficient alternative by reconstructing rare but important traffic scenarios. More importantly, it allows simulating risky scenarios that are usually difficult to obtain in real-world driving.

Traditional simulators often rely on static log replay that simulates the agent behavior based on what happened in the collected data. It fails to produce reactive behavior of environment agents when the ego plan diverges from the original log and thus becomes less useful in interactive scenarios. On the other hand, heuristic-based models, such as the intelligent driver model (IDM)~\cite{treiber2000congested, liu2021deep,huang2021tip}, produce more reactive behaviors in response to diverging ego plans, but they are limited to following predefined trajectories and have difficulties in producing diverse scenarios. 

Recent advances in machine learning have enabled realistic and diverse agent simulation by training agent models from realistic driving data. They demonstrated great potential in supporting closed loop planner evaluation. For instance, \cite{bergamini2021simnet} proposes to generate realistic driving episodes by leveraging a probabilistic prediction model given traffic observations and environmental context; \cite{sun2022rtgnn} infers future agent states as both discrete intent and continuous controls conditioned on past observations over all agents. On the other hand, they focus on simulating trajectories of individual agents without reasoning about their future interactions, which could lead to colliding trajectories in dense traffic. 
To overcome this challenge,~\cite{zhou2021exploring} adds a task loss to penalize collisions and~\cite{vitelli2021safetynet} proposes a feasibility check on the generated trajectories to filter out colliding trajectories. Instead of requiring a hand-crafted loss or an ad-hoc filter, \cite{suo2021trafficsim} offers simulation consistency by rolling outs joint trajectories over all the agents in a scene through an implicit latent variable learned by a conditional variational autoencoder; however, such generative models rely on probabilistic sampling and suffer from producing rare or dangerous scenarios, which are crucial to testing autonomous driving planners. 

\begin{figure}[t!]
\vspace{2mm}
    \centering
    \includegraphics[width=0.47\textwidth]{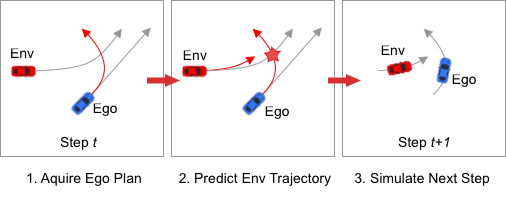}
    \caption{A motivating example of \modelname. Given an updated ego plan in red, \modelnamespace first identifies the relevant agent whose future trajectories may collide with the ego plan, and infers the interacting relations between the agents, such as the environment agent is yielding to the ego agent. It then predicts consistent and reactive trajectories for the relevant agent given the inferred relations. The simulator simulates the next step of all agents based on the predictions and waits for the new plan.}
    \label{fig:stepping_example}
\vspace{-4mm}
\end{figure}

In this work, we propose \modelname, an interactive traffic simulation model that reasons about explicit interaction relations when rolling out future trajectories for all agents in a scene. As shown in Fig.~\ref{fig:stepping_example}, given an updated ego plan in red, \modelnamespace first identifies the relevant agents whose future trajectory may collide with the new ego plan. Next, it infers the interaction relations between the ego agent and the relevant environment agents, and predicts reactive trajectories for the relevant agents based on the relation. Finally, it follows the predictions to simulate one step ahead for the environment agents and repeat the same procedure until the end of the simulation episode.

Compared to existing learning-based models, \modelnamespace offers a few advantages by reasoning about the explicit interactive relations among the relevant traffic agents. 
First, the relations guide the trajectory simulator to produce consistent trajectories of multiple agents in complex interactive scenarios and offer better interpretability. 
Second, one can use it to manipulate an interactive scenario by specifying the interaction relation between agents, which is a non-trivial task for generative models that may require a large number of samples. 
Third, it affords better efficiency by identifying only the relevant agents influenced by the ego plans and modifying their future trajectories in simulation as opposed to all agents.

Our contribution is as follows:
\begin{itemize}
\item We propose a learning-based simulation model, \modelname, that rolls out realistic and consistent future trajectories of multiple traffic agents based on explicit interaction relations.
\item We leverage a relation predictor to infer interaction relations for better interpretability and simulation efficiency, and show how our simulator can be used to manipulate different interaction situations by specifying the relations.
\item We train and evaluate our model on the Waymo Open Motion Dataset, a publicly available real-world driving benchmark, and demonstrate its advantage compared to a state-of-the-art baseline in two simulation tasks.
\end{itemize}





\section{Related Work}
In this section, we discuss relevant literature in three aspects: traffic simulation, behavior prediction, and interaction modeling.

\subsection{Traffic Simulation}
Traffic simulation is an important task for intelligent transportation systems, allowing for training and evaluating driving models in a more scalable and safe way. Existing traffic simulators render high-fidelity driving environments in the context of racing~\cite{wymann2000torcs} and urban driving~\cite{dosovitskiy2017carla,lopez2018microscopic}. However, they often simulate agent behaviors through heuristic-based models that fail to cover diverse scenarios or interactions. 

Recently, learning-based models have demonstrated great success in simulating realistic and reactive agent behaviors by learning driving patterns from real-world driving data. For instance, \cite{bergamini2021simnet} trains a deep neural network through a rasterized representation derived from driving logs to simulate future agent trajectories; \cite{sun2022rtgnn} infers future agent intent and control inputs to model stochastic traffic dynamics. While such methods consider the past trajectories of all the agents at once, they assume independence of future trajectory rollouts that may lead to inconsistent or colliding trajectories between simulated agents in interactive scenarios. 

In order to improve simulation consistency over multiple interacting agents,~\cite{zhou2021exploring} leverages a collision loss and~\cite{vitelli2021safetynet} proposes a rule-based fall-back layer to discourage or avoid collisions. While such works often require hand-crafted losses or post-processing filters,~\cite{suo2021trafficsim} proposes a multi-agent behavior model that simulates joint agent behaviors directly through an implicit latent variable that governs the agent interactions. Compared to existing models, we propose a relation-aware simulator that simulates diverse and realistic interactive behaviors in a more straightforward and efficient way by explicitly modeling interacting relations.

\subsection{Behavior Prediction}
Behavior prediction offers a natural solution to simulate agent behaviors through the predicted trajectories given the environmental context. Recent models prove great success in improving prediction accuracy, by learning agent dynamics and environmental context represented either as a vector representation~\cite{gao2020vectornet,liang2020learning} or a rasterized image~\cite{cui2019multimodal,gilles2021home}.

Due to uncertainty in human intent, the future trajectories are multi-modal. To handle the multi-modality and improve prediction coverage, a family of models are proposed to first predict high-level intent, such as goal targets  \cite{rhinehart2019precog,zhao2020tnt,mangalam2020pecnet}, lanes to follow~\cite{song2021learning,kim2021lapred}, maneuvers \cite{deo2018multi,huang2020diversitygan,huang2021hyper}, and linguistic descriptions~\cite{kuo2022trajectory}, before predicting low-level trajectories that are conditioned on the intent. 

In this work, we take advantage of the goal-conditioned models in the behavior prediction literature to simulate realistic agent trajectories given the environmental context and agent intent.


\subsection{Interaction Modeling}
Modeling interaction is an important task in motion prediction and simulation when reasoning about multi-agent behaviors. While many existing approaches \cite{casas2020implicit,chen2020hgcn,casas2020spagnn,kamra2020mapred2} rely on implicit latent variables to model interactions, we focus on modeling and predicting explicit interaction relations in this work for better interpretability. These explicit relations allow us to produce and manipulate different types of interactive scenarios.

In this work, we follow \cite{lee2019joint,kumar2020interaction,sun2022m2i} that define agent relations based on the pass and yield relationship and predict the relationship as a classification problem through a separate learning model. The predicted relations are useful in guiding the motion predictor to generate consistent trajectories among multiple agents, as shown by~\cite{sun2022m2i}.

When there exist potential conflicts between a novel ego plan
and the simulated trajectories of environment agents given the predicted relations, we adopt conflict resolution techniques that are widely used in planning~\cite{adler1989conflict}, search~\cite{moskewicz2001chaff}, and ordering~\cite{chen2019efficiently}. 

\begin{figure*}[t!]
\vspace{2mm}
    \centering
    \includegraphics[width=1\textwidth]{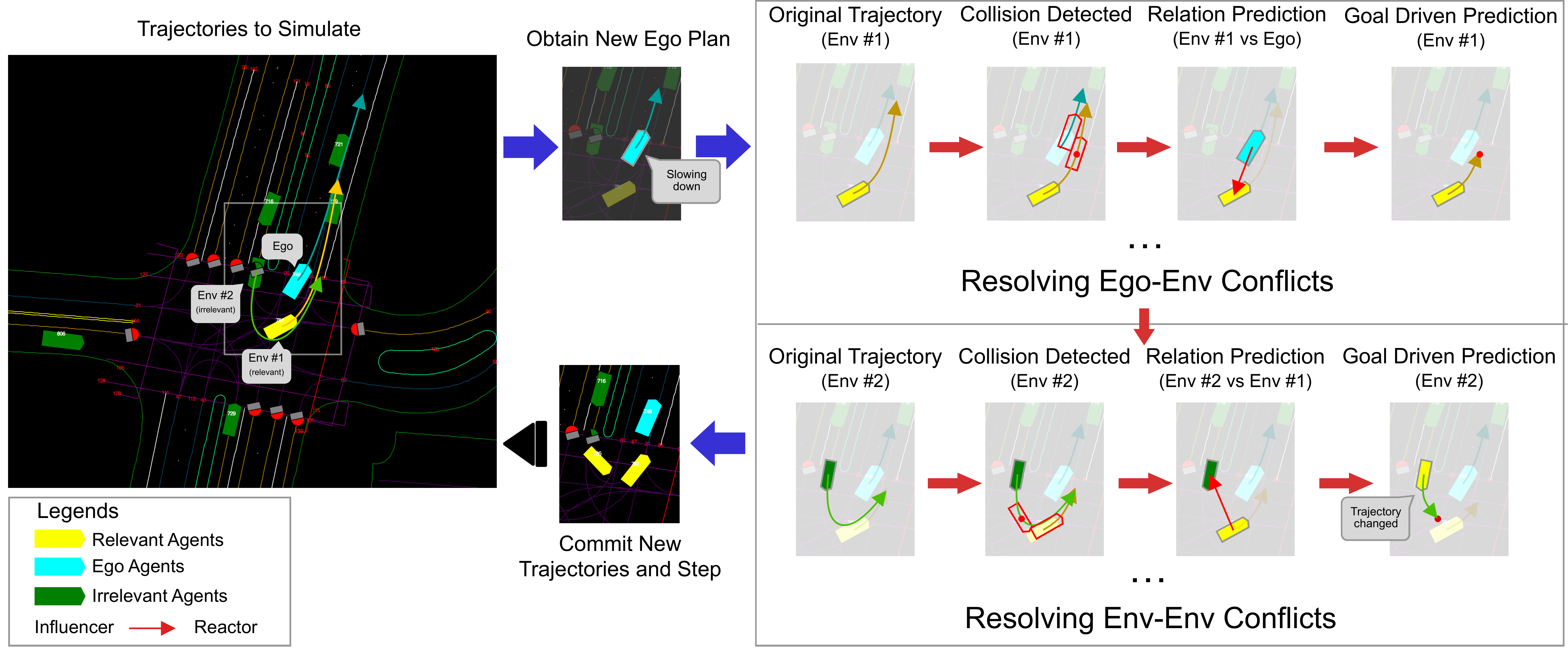}
    \caption{Illustration of \modelname. In this example, given a new plan for the ego agent (in cyan) to slow down, the simulator updates its simulated trajectories for the environment agents as follows. First, it checks for potential collisions with all environment agents and labels colliding ones as the relevant agents in yellow$^1$. For each relevant agent, such as Env \#1, it predicts the interaction relation and updates its trajectory based on the relation using a goal driven trajectory predictor. Second, it resolves collisions between the newly updated trajectories of the environment agent(s) and the remaining agents (i.e. Env \#2) iteratively until all collisions are resolved. In the end, \modelnamespace successfully generates scene consistent trajectories for Env \#1 and Env \#2 to react and slow down, and commit these trajectories to simulate for the next step.}
    \label{fig:collisions}
\vspace{-6mm}
\end{figure*}

\section{Problem Formulation}
We formulate the problem of learning realistic interactive behaviors for traffic simulation following~\cite{suo2021trafficsim}. Given map states $M$ and the observed states $S$ of $N$ traffic agents in a scene, the goal is to roll out the future states of all agents $Y$ up to a finite horizon $T$.

Due to the computational complexity and memory constraint in simulating joint behaviors over all traffic agents in the scene, our model focuses on simulating agent behaviors that are \emph{relevant} to the ego plan, as the irrelevant agent behaviors are often ignored by the ego planner. 
For an irrelevant agent whose future trajectory stays the same given a new ego plan at the next step, our simulator can simply roll out its future trajectory from the data.

One key consideration in our problem is to faithfully follow the agent's origin intent as much as possible. We define such intent based on the goal location collected from ground truth future trajectories in the data. Although this consideration may suffer from less diverse scenarios, we observe that most drivers do not change their long-term goals when interacting with others (e.g., an agent turning left in the data is unlikely to change its intent to go straight or turn right) and we can achieve sufficient diversity by simulating multiple options to reach the long-term goals, as we show in the experiments. 


\section{Approach}
In this section, we describe our approach \modelnamespace that simulates interactive responses of environment agents given an ego plan to be tested. We first present a high-level overview with an illustrated example followed by detailed explanations on conflict detection, relation-aware conflict resolution, goal driven trajectory prediction, conflict resolution between environment agents, and one-step simulation. 
\stepcounter{footnote}\footnotetext{The simulator has already labeled some relevant agents in yellow from the previous simulation step. In the current step, it will re-evaluate the relevance for all agents.}


\subsection{Overview}
As illustrated in Fig.~\ref{fig:collisions}, \modelnamespace takes the input, including the past agent states and the environmental context, into a rasterized representation and a vectorized representation. Given a new ego plan (in cyan), \modelnamespace identifies the potential conflicts defined as the collisions between the plan and the simulated trajectory of other agents, as shown on the first row in Fig.~\ref{fig:collisions}. For the conflicting agents, \modelnamespace predicts their interaction relations with the ego agent and updates their future trajectories or the ego plan conditioned on the predicted relations and their intent. In cases where the updated trajectories cause new conflicts with the other environment agents, \modelnamespace further updates the colliding trajectories of these agents using the same relation prediction and trajectory prediction procedure until all conflicts are resolved, as shown on the second row in Fig.~\ref{fig:collisions}. 
Following this tree-like structure to resolve collisions, \modelnamespace is guaranteed to produce no dead-end cases. After updating the trajectories of all the relevant environment agents, \modelnamespace advances the simulation by one step and repeats this process until the end of the simulation episode. 

\subsection{Conflict Detection}
\label{sec:conflict_detection}
At each simulation step, the ego planner supplies a planned trajectory that may result in conflicts with the future trajectories of the environment agents from log replay or past simulations. The conflict is defined as the collision between agents based on their shapes represented as bounding boxes, as customary in planning~\cite{huang2021risk}. If the bounding boxes between two agents overlap at any time in the future, a conflict is detected and requires the simulator to update the colliding trajectories for better consistency and realism. We present an example of conflict highlighted by the red boxes on the top row (third to the right plot) in Fig.~\ref{fig:collisions}, which indicate the collision between the ego agent and environment agent \#1.

\subsection{Relation-Aware Conflict Resolution}
\label{sec:relation_pred}
To resolve a collision conflict, \modelnamespace first identifies the \emph{relevant} agents that have colliding trajectories with the ego plan and revises their trajectories or the ego plan based on the predicted relations between the ego and relevant agents.

In order to model the interaction relation, we define the interactive agents as the influencer and the reactor, as customary in~\cite{kumar2020interaction,sun2022m2i}, and predict the relation as a classification problem through a deep neural network. The network includes a context encoder that encodes the input using two encoder heads. The first encoder head follows VectorNet~\cite{gao2020vectornet} by taking observed agent trajectories and map states as a set of polylines. For each polyline, the encoder first runs an MLP to encode the feature of each vector within the polyline and a graph neural network followed by a max-pooling layer to extract the polyline feature from accumulated vector features. The final vectorized feature is obtained by running cross attentions between the agent polyline features and the map polyline features. Due to the success of rasterized representations in trajectory prediction~\cite{gilles2021home}, we leverage a second encoder head by rasterizing the input states into an image with multiple channels that include the position of the agents at each past time step with the map information and obtaining the encoded rasterized feature through a pre-trained VGG16~\cite{simonyan2014very} model. The output of the context encoder is a concatenation of the vectorized feature and the rasterized feature from both encoder heads. Finally, the network predicts the probability over interaction relations using a two-layer MLP. The relation predictor is trained using a cross entropy loss between the predicted probabilities and the ground truth interaction label. The label is obtained at training time based on which agent gets to the cross point first given their ground truth future trajectories in the data, as in~\cite{sun2022m2i}.

After training a relation predictor from realistic driving data, we can resolve the collision conflict based on the definition that the reactor is always yielding to the influencer. Thus, we keep the trajectory of the influencer unchanged and modify the goal of the reactor to avoid colliding with the influencer trajectory. There exist a few options to set the goal for the reactor, depending on the trade-off between progress (i.e. how far the reactor travels) and simulation consistency (i.e. how likely the agents are colliding). In this work, we update the goal point of the reactor to the colliding point (or the crossing point) between the trajectories as shown in the top right plot in Fig.~\ref{fig:collisions} to favor simulation consistency, such that the agents are unlikely to collide, while ensuring the progress to some extent.

\subsection{Goal Driven Trajectory Prediction}
We train a goal-driven trajectory predictor as a deep neural network to help simulate realistic reactor trajectories given its updated goal at the cross point. This predictor is also used to roll out the agent trajectories from novel states unseen in the dataset, or update the ego plan if the ego agent is predicted as the reactor. This network shares the same context encoders as the relation predictor, except that it adds an additional channel in the rasterized representation to include the goal point. The decoder is an MLP-based predictor head that regresses the full trajectory conditioned on the goal. At training time, we use the teacher forcing technique and provide the ground truth goal point from the data. We train the model to minimize the L2 loss between the regressed trajectory and the ground truth future trajectory from data.

\subsection{Conflict Resolution between Environment Agents}
\label{sec:conflict_res_env}
After modifying the trajectories to resolve the conflict between the ego and its relevant agents, it is possible that new collisions are created by the modified trajectories. Thus, we follow the same conflict detection and resolution procedure iteratively for agents whose future trajectories are colliding with the updated trajectories until all conflicts are resolved. In the example illustrated in Fig.~\ref{fig:collisions}, after \modelnamespace updates the trajectory of environment agent \#1 to resolve its collision with the new ego plan (see the top row), it detects a new collision between agent \#2 and the updated trajectory of agent \#1. In this case, it continues to resolve this conflict by predicting the relations between these two agents and updating the future trajectory of agent \#2 as the reactor (see the second row). Following this procedure, our simulator will keep searching for the collisions given newly updated trajectories until all conflicts are resolved. In the end, it produces scene consistent trajectories over all agents.

\subsection{Simulation}
Given the predicted trajectories of all agents in a scene, \modelnamespace commits these trajectories to simulate the agent behaviors at the next step. If the plan stays the same at the next step, \modelnamespace will follow the remaining predicted trajectories at each future step, until the end of the simulation horizon. On the other hand, if a new plan is given, it will follow the procedures described in Sec.~\ref{sec:conflict_detection} to Sec.~\ref{sec:conflict_res_env} to update the environment agent trajectories when necessary.

\section{Experimental Results}
\label{sec:result}
In this section, we introduce the dataset and the model, and present a series of quantitative and qualitative experiments to demonstrate the effectiveness of our approach compared to the baseline simulators.

\begin{table*}[t!]
    \centering
    \footnotesize
    \bgroup
    \def\arraystretch{1.05}%
    \begin{tabular}{lccccccc}
    \toprule
    Method & Relevant Ratio & ADE $\downarrow$ & FDE $\downarrow$  & Front Collision Rate $\downarrow$ & Side Collision Rate $\downarrow$ & Rear Collision Rate $\downarrow$ & Progress $\uparrow$ \\
    \midrule
    SimNet~\cite{bergamini2021simnet} & 21\% & 8.74 & 16.99 & 0.89\% & 2.80\% & 4.97\% & 49.35 \\ 
    \modelname-M0 & 26\% & 7.26 & 6.19 & 0.57\% & 1.99\% & 9.28\% & \textbf{54.11} \\
    \modelname-M1 & 23\% & 8.31 & 6.58 & \textbf{0.10\%} & \textbf{0.41\%} & \textbf{1.00\%} & 39.83 \\
    \modelname & 23\% & \textbf{7.11} & \textbf{6.17} & 0.19\% & 0.71\% & 1.91\% & 42.85 \\
    \bottomrule
    \end{tabular}%
    \egroup
    \caption{Task 1 simulation performance (c.f. Sec.~\ref{sec:quant} for more details on the metrics) of \modelnamespace and the baselines over 40k interactive scenarios, during which the ego agent is controlled by a state-of-the-art marginal predictor~\cite{gu2021densetnt}. Compared to the baselines, \modelnamespace achieves the best simulation realism at the cost of small simulation reactivity by modeling explicit interaction relations during simulation.}
    \label{tab:task1}
\vspace{-6mm}
\end{table*}

\begin{table}[t!]
    \centering
    \footnotesize
    \bgroup
    \def\arraystretch{1.05}%
    \begin{tabular}{lccccc}
    \toprule
    Method & Rel. Rat. & Front $\downarrow$ & Side $\downarrow$ & Rear $\downarrow$ & Prog. $\uparrow$ \\
    \midrule
    SimNet~\cite{bergamini2021simnet} & 9.33\% & 0.60\% & 4.55\% & 33.90\% & 46.55 \\
    \modelname-M0 & 11.06\% & 0.47\% & 4.90\% & 35.00\% & \textbf{50.69} \\
    \modelname-M1 & 10.38\% & \textbf{0.24\%} & 1.98\% & \textbf{6.94\%} & 26.67 \\
    \modelname & 10.49\% & 0.30\% & \textbf{1.95\%} & 7.48\% & 28.47 \\
    \bottomrule
    \end{tabular}%
    \egroup
    \caption{Task 2 simulation performance of \modelnamespace and the baselines over 40k interactive scenarios, during which the ego agent follows a slowing down action, requiring the simulator to generate reactive behaviors to avoid collisions. Results in this more challenging task show that an explicit collision resolution process helps decrease the collision rate for all types of collisions by a large margin.}
    \label{tab:task2}
\vspace{-6mm}
\end{table}

\subsection{Dataset and Model Details}
We train our simulator on the Waymo Open Motion Dataset (WOMD) \cite{ettinger2021large}, a large-scale driving dataset collected from realistic traffic scenarios. More specifically, we focus on the interactive dataset in WOMD that is mined to involve close interactions. At test time, we evaluate our simulator against the validation set, which includes $42,318$ challenging scenarios. In each scenario, the simulator takes as the input one second of the observed trajectory and the environmental context, and simulates the agent behaviors over the next 8 seconds. We follow~\cite{suo2021trafficsim} and choose a simulation frequency at 2Hz.

We implement our relation predictor and goal driven trajectory predictor by following~\cite{sun2022m2i}, a state-of-the-art interactive trajectory predictor on the WOMD interactive prediction benchmark. Both predictors leverage the same context encoder headers that encode the observed trajectories and environment states into a vectorized representation and a rasterized representation. More specifically, we implement a standard VectorNet encoder~\cite{gao2020vectornet} following the open-source implementations by~\cite{gu2021densetnt}. To obtain the rasterized feature, we generate a $224 \times 224$ image from the past trajectories and the environmental context, in which each pixel represents an area of $1\mathrm{m} \times 1\mathrm{m}$, and run a pre-trained VGG16~\cite{simonyan2014very} model as the raster encoder. The relation prediction head is a two-layer MLP. The first layer has a hidden size of 128, followed by a layer normalization layer and a ReLU activation layer. The second layer outputs the distribution over the binary interaction relations. The goal driven prediction head follows a similar two-layer MLP, consisting of a hidden layer with a size of 128, a layer normalization layer, a ReLU activation layer, and an output layer that outputs the 2-D positions over the simulation horizon. We refer to~\cite{gu2021densetnt,sun2022m2i} for additional implementation details.

Both models are trained on the training set from WOMD with a batch size of 64 for 30 epochs. We use an Adam optimizer and a learning rate scheduler that decays the learning rate by 30\% every 5 epochs, with an initial value of 1e-3.

We further introduce a few baselines to verify the effectiveness of our approach, which include 1) \emph{SimNet}~\cite{bergamini2021simnet} that simulates agent trajectories using a ResNet-50 model without reasoning about their future interactions; 2) \emph{\modelname-M0} is a variant of our model that simulates agent trajectories but does not resolve the collision conflicts; 3) \emph{\modelname-M1} that is a variant of our model that resolves the collision conflict in a conservative manner by simulating both agents to the cross point without considering their relations, which shares the same spirit as~\cite{vitelli2021safetynet}. For each variant, we use the same backbone as our model for a fair comparison.

\subsection{Simulation Tasks}
We propose two tasks to test the effectiveness of our simulator. In the first task, we test its capability to generate realistic and consistent trajectories given a proper planner. In the second task, we test its stability against a planner that generates novel out-of-distribution plans with respect to the WOMD dataset. In both tasks, we randomly select the ego agent to be one of the interactive agents labeled in the dataset to test our simulator in highly interactive scenarios.

\subsubsection{Task 1: Simulation with Respect to Proper Plans}
To test the overall performance of \modelname, we control the ego agent using a state-of-the-art marginal trajectory predictor~\cite{gu2021densetnt}, which was the winner of the WOMD marginal prediction challenge. This allows us to simulate realistic ego behaviors that closely follow the data distribution but deviate slightly from the ego future trajectory collected in the data. As a result, it is necessary for the simulator to adjust future trajectories of the environment agents to react to the deviated plan.

\subsubsection{Task 2: Simulation with Respect to Novel Plans}
Learning-based models often suffer from instability given novel inputs, such as the ego trajectories that are out of the distribution of the training dataset. Therefore, one should verify the stability of the simulation given novel ego plans. In this task, we control the ego vehicle to commit a slowing down action with a reasonable deceleration in each scene to test the robustness of our simulator and the baselines, which are required to produce reactive behaviors for the environment agents, especially those behind the ego agent. 
The deceleration is set to $-1.5 m/s^2$. We upper bound the speed of the ego plan by its original speed from the dataset, in case the ego vehicle is going to slow down with a larger deceleration.
As the slowing down action is easy to implement and reproduce, we find it straightforward to test and compare the performance of interactive simulators. 

\begin{figure*}[t!]
    \centering
    \includegraphics[width=1\textwidth]{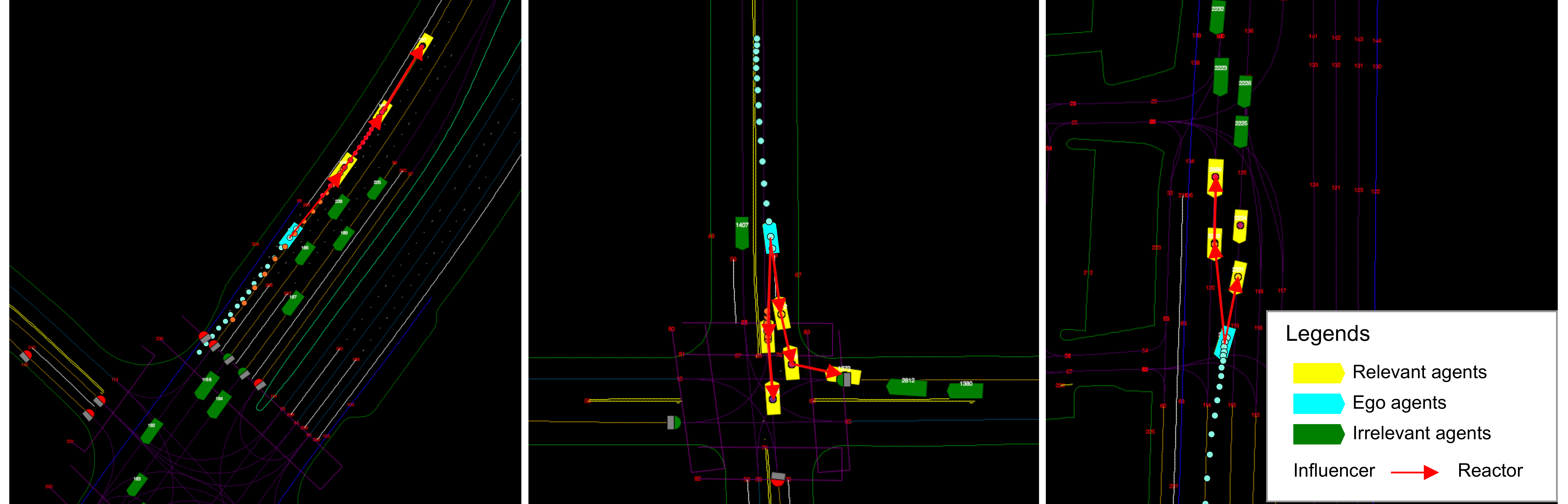}
    \caption{Examples of \modelnamespace successfully identifying relevant agents in challenging interactive scenarios and simulating realistic interactive behaviors. In the first scenario from the left, our model identifies all the platooning agents in yellow that are relevant to the ego agent plan. In the second and third scenarios, our model identifies the agents with complex interactions during turning and merging. The planned ego trajectories and the simulated trajectories for each relevant environment agent are drawn as dots in different colors.}
    \label{fig:show_case}
\vspace{-6mm}
\end{figure*}

\subsection{Quantitative Results}
\label{sec:quant}
We evaluate the quantitative performance of our model and the baselines against a few standard metrics introduced in~\cite{bergamini2021simnet,suo2021trafficsim,caesar2021nuplan}, including: 
\begin{itemize}
    \item \textbf{Relevant ratio} as the number of relevant agents whose future trajectories are modified by the simulator over the number of all environment agents. This measures the density of agents that needs to be simulated in WOMD.
    \item \textbf{Average displacement error (ADE)} and \textbf{final displacement error (FDE)} as the distance between the simulated trajectories and the ground truth future trajectories from data. These two metrics measure the simulation realism~\cite{bergamini2021simnet} and the scenario reconstruction ability~\cite{suo2021trafficsim}.
    \item \textbf{Collision rate} as the number of the colliding agent pairs divided by the number of simulated agents. We follow~\cite{bergamini2021simnet} and divide the collisions into front collisions, side collisions, and rear collisions. Such metric measures the simulation reactivity~\cite{bergamini2021simnet} and the interaction reasoning capability~\cite{suo2021trafficsim}.
    \item \textbf{Progress} as the total traveled distance of all simulated agents divided by the number of simulated agents. This measures the level of goal achievement~\cite{caesar2021nuplan}.
\end{itemize}
The results are reported in Table~\ref{tab:task1} and Table~\ref{tab:task2} for Task 1 and Task 2, respectively. We summarize the key observations as follows.

\subsubsection{Simulation Efficiency}
In both simulation tasks, we observe a small portion (i.e. at most 26\% for Task 1 and at most 11.06\% for Task 2) of relevant agents that are influenced by the ego plan out of all environment agents. This demonstrates that a simulator only needs to modify a small portion of agents on the road to test a planner's performance to achieve better run-time efficiency. A joint trajectory simulator such as~\cite{suo2021trafficsim}, on the other hand, might not be the most efficient method especially under time constraints or memory limits.

\subsubsection{Simulation Realism}
In Task 1, we measure the simulation realism as the displacement errors, i.e. ADE and FDE, between the simulated trajectories and the ground truth future trajectories in the data, and observe that our simulator achieves the lowest errors, especially compared to SimNet that leverages a simple ResNet-50 prediction backbone. When compared to the other two invariants using the same prediction backbone, \modelnamespace achieves the best performance by reasoning about the underlying relations between the agents.

\subsubsection{Simulation Reactivity}
We measure the simulation reactivity in terms of collision rates in both tasks. In Task 1 (see Table~\ref{tab:task1}), while both \modelname-M1 and \modelnamespace achieve better reactivity by reasoning about future interactions, \modelname-M1 achieves the lowest collision rates as it adopts a conservative approach to resolve conflicts by changing the goals of both interactive agents to the cross point. On the other hand, our approach \modelnamespace resolves the conflicts by only modifying the reactor agent and achieves a better balance between simulation realism and reactivity.

We observe a similar pattern in Task 2 (see Table~\ref{tab:task2}), in which we test the simulator with a slowing down ego plan. Both SimNet and \modelname-M0 struggle to resolve new conflicts and cause rear collisions in one-third of the scenarios by simulating the rear agent to collide with the slowing down ego agent. Despite having the lowest values in two types of collisions, \modelname-M1 handles conflicts by yielding both agents without really reasoning about their interactions. On the other hand, our proposed approach \modelnamespace makes more progress at similar collision rate levels by resolving conflicts via explicit interaction modeling.

\begin{figure*}[t!]
    \centering
    \includegraphics[width=0.9\textwidth,trim={0 0 0 1cm},clip]{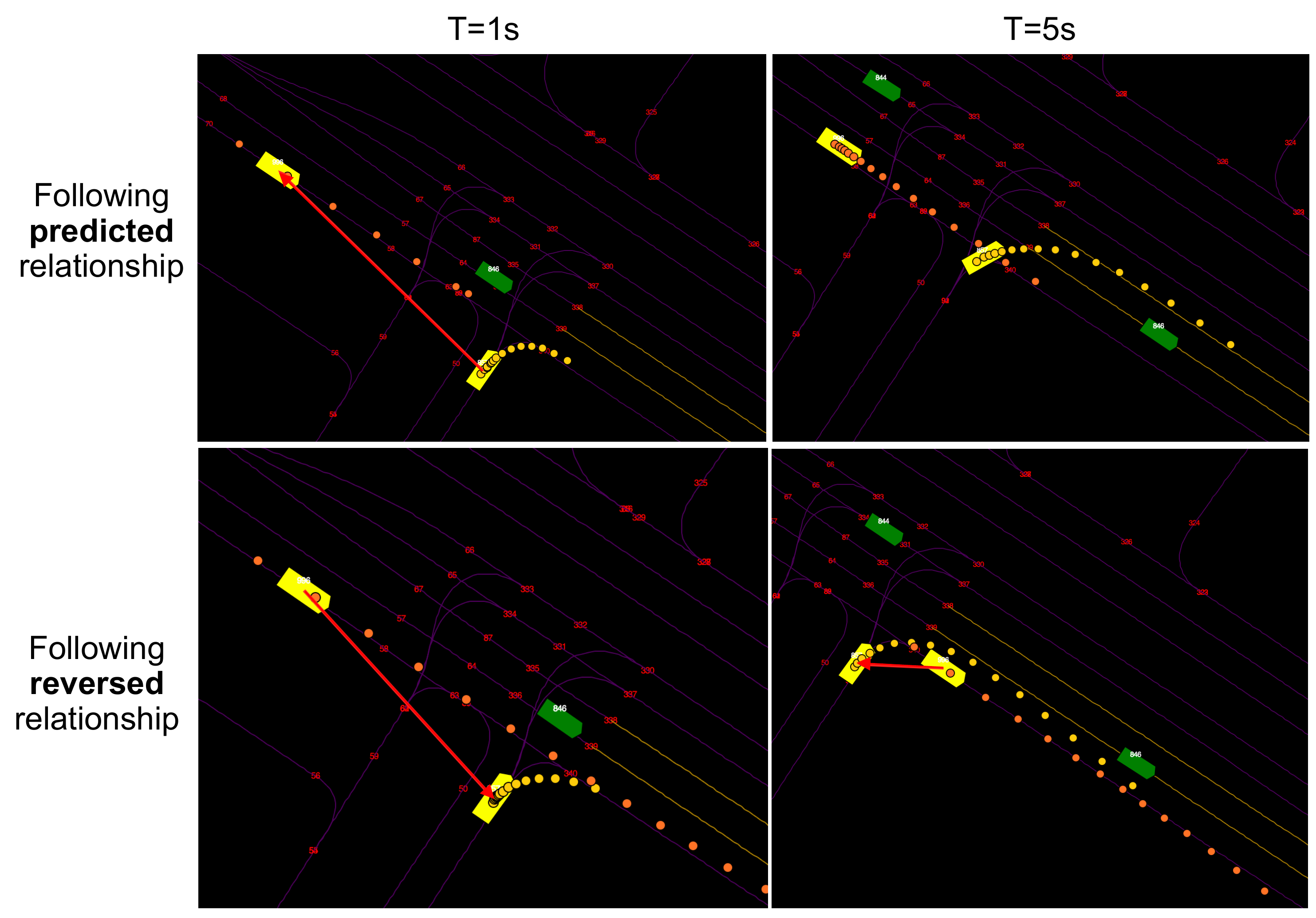}
    \caption{An example of \modelnamespace manipulating the interaction relationship between two agents to simulate different interaction modalities. Top row: The model follows the predicted relation and simulates the forward agent to yield to the turning agent who has already initiated the turn. Bottom row: The model follows a reversed relation to simulate a different scenario.}
    \label{fig:relation_perturb}
\vspace{-6mm}
\end{figure*}
\subsubsection{Simulation Progress}
We measure the level of goal achievement for all the agents through the progress metric. A good planner should control the ego agent to navigate safely without blocking other traffic. In both tasks, we observe that our approach achieves better progress compared to \modelname-M1 that resolves the interaction conflicts conservatively. In addition, we notice that the simulators tend to achieve lower progress scores in Task 2, due to the slowing down ego behavior.

\subsection{Qualitative Examples}

\subsubsection{Realistic Behavior Simulation}
In Fig.~\ref{fig:show_case}, we present three representative interactive scenarios from Task 1 to showcase \modelname. Our model successfully identifies all relevant agents, as highlighted in yellow, in challenging scenarios involving platooning, turning, and merging, and simulates their future trajectories to be consistent with the interaction relations and the environmental context.
 
\subsubsection{Interaction Manipulation with Perturbed Relations}
In Fig.~\ref{fig:relation_perturb}, we present an example in which \modelnamespace can be used to manipulate the interaction relations between agents to simulate different interactive behaviors. The simulator in this example is used to simulate the trajectories of a forward agent and a turning agent. On the top row, our relation predictor predicts that the forward agent is yielding to the turning agent who already initializes the turn. Given the predicted relation, \modelnamespace simulates a realistic interacting scenario for both agents. In contrast, we can reverse the interaction relation and simulate a different scenario, in which the turning agent yields to the forward agent, as shown on the bottom row. This demonstrates the advantage of our approach that manipulates relations and simulates diverse and rare events, compared to generative models, i.e.~\cite{suo2021trafficsim}, that rely on probabilistic sampling approaches and may require a large number of samples to cover rare events. 

\subsection{Collisions Analysis}
While our model is designed to resolve all collision conflicts in simulation, we observe that the collision rates in the experimental results are not eliminated to zero. We present a couple of reasons behind such failure cases.

First, the goal driven trajectory prediction model is not always stable and may fail given novel inputs, such as strange observations from other agents. Such novel inputs often cause the trajectory to overshoot and lead to a rear collision. 
Second, due to noise in perception data, including the bounding box labels and the agent positions, there exist a few scenarios that already include collisions at the beginning of the simulation.



\section{Conclusions}
In conclusion, we present an interactive traffic simulator \modelnamespace that simulates realistic and reactive agent trajectories with respect to the ego plan. 
Our simulator takes advantage of a relation predictor to reason about the underlying relations between interactive agents when simulating their future trajectories.
In the experiments, we test our simulator and the baselines with two planners, a reasonable planner and a novel planner, to validate the advantage of our approach. Quantitative results show that \modelnamespace achieves a good balance between simulation realism and reactivity, especially in a challenging task in which the ego plan is forced to slow down and requires the simulator to respond. We also present a few representative examples to demonstrate the effectiveness of \modelnamespace in dense interactive scenarios and its ability to manipulate agent interactions to produce diverse and rare events.



\bibliographystyle{IEEEtran}
\bibliography{references} 

\end{document}